\documentclass[11pt]{article}

\PassOptionsToPackage{table}{xcolor}
\usepackage[preprint]{acl}
\usepackage{times}
\usepackage{latexsym}
\usepackage[T1]{fontenc}
\usepackage[utf8]{inputenc}
\usepackage{microtype}
\usepackage{inconsolata}

\usepackage{amsmath,amssymb,mathtools}
\usepackage{booktabs,graphicx,array,tabularx,multirow}
\usepackage{placeins}

\usepackage{etoolbox}
\newcommand{\displaymathfontsize}{\fontsize{10pt}{\baselineskip}\selectfont}
\AtBeginEnvironment{equation}{\displaymathfontsize}
\AtBeginEnvironment{equation*}{\displaymathfontsize}
\AtBeginEnvironment{align}{\displaymathfontsize}
\AtBeginEnvironment{align*}{\displaymathfontsize}
\AtBeginEnvironment{alignat}{\displaymathfontsize}
\AtBeginEnvironment{alignat*}{\displaymathfontsize}
\AtBeginEnvironment{flalign}{\displaymathfontsize}
\AtBeginEnvironment{flalign*}{\displaymathfontsize}
\AtBeginEnvironment{gather}{\displaymathfontsize}
\AtBeginEnvironment{gather*}{\displaymathfontsize}
\AtBeginEnvironment{multline}{\displaymathfontsize}
\AtBeginEnvironment{multline*}{\displaymathfontsize}
\AtBeginEnvironment{displaymath}{\displaymathfontsize}
\AtBeginEnvironment{eqnarray}{\displaymathfontsize}
\AtBeginEnvironment{eqnarray*}{\displaymathfontsize}

\newcolumntype{R}{>{\raggedleft\arraybackslash}X}
\newcolumntype{Y}{>{\hsize=1.8\hsize\raggedright\arraybackslash}X}
\newcolumntype{N}{>{\hsize=0.8\hsize\raggedleft\arraybackslash}X}

\DeclareMathOperator*{\argmax}{arg\,max}
\DeclareMathOperator*{\argmin}{arg\,min}
\newtheorem{theoryproposition}{Proposition}

\newcommand{\ours}{D\textsuperscript{2}-OPD}

\definecolor{tableNavy}{HTML}{274961}
\definecolor{tableMuted}{HTML}{59646B}
\definecolor{tableRule}{HTML}{BBC5CB}
\definecolor{tableNegative}{HTML}{8E534E}
\definecolor{summaryband}{HTML}{EEF3F6}
\definecolor{oursband}{HTML}{FAEEE9}
\definecolor{ourssummary}{HTML}{F2DAD0}
\definecolor{oursink}{HTML}{9B4538}
\definecolor{refband}{HTML}{F2F2F2}

\newcommand{\best}[1]{\textbf{#1}}
\newcommand{\gain}[1]{{\small\textcolor{oursink}{#1}}}

\title{\ours{}: Separating Direction and Distance in Extrapolative \\ On-Policy Distillation}
\author{
  Bowen Fang\thanks{This work was done during an internship at Amazon.} \\
  Columbia University \\
  \texttt{bf2504@columbia.edu}
  \And
  Dingmin Wang \\
  AWS AI Lab \\
  \texttt{wdimmy@amazon.com}
  \And
  Hengzhi Pei \\
  Amazon AGI \\
  \texttt{philepei@amazon.com}
}

\date{}
\hypersetup{
  pdftitle={D2-OPD: Separating Direction and Distance in Extrapolative On-Policy Distillation},
  pdfauthor={Bowen Fang, Dingmin Wang, Hengzhi Pei}
}

\begin{document}
\begin{NoHyper}
\maketitle
\end{NoHyper}

\begin{abstract}

On-policy distillation (OPD) trains a student model to match a teacher's next-token distributions on student-generated prefixes.
Extrapolative OPD further extends the teacher's log-probability shift relative to a reference policy to construct stronger learning targets. However, existing methods couple extrapolation direction and distance, allowing the magnitude of the teacher's shift to dictate how far the target moves. 
We propose \ours{}, decoupling these two factors: a per-prefix hard KL constraint bounds the target’s distance from the teacher, while an adaptive combination of the teacher’s post-training shift and the initial student–teacher gap determines the extrapolation direction.
We optimize the student toward this target using forward KL, providing learning signals even for low-probability tokens.
We evaluate three settings: where standard OPD performs well, where it struggles with base-model initialization, and where no teacher reference is available.
On mathematical reasoning benchmarks, \ours{} attains the highest score in all three settings, outperforming the strongest baseline by up to 9.45 points. 
Ablations validate the contributions of each component.
\end{abstract}

\begin{figure*}[tp]
\centering
\includegraphics[width=\textwidth]{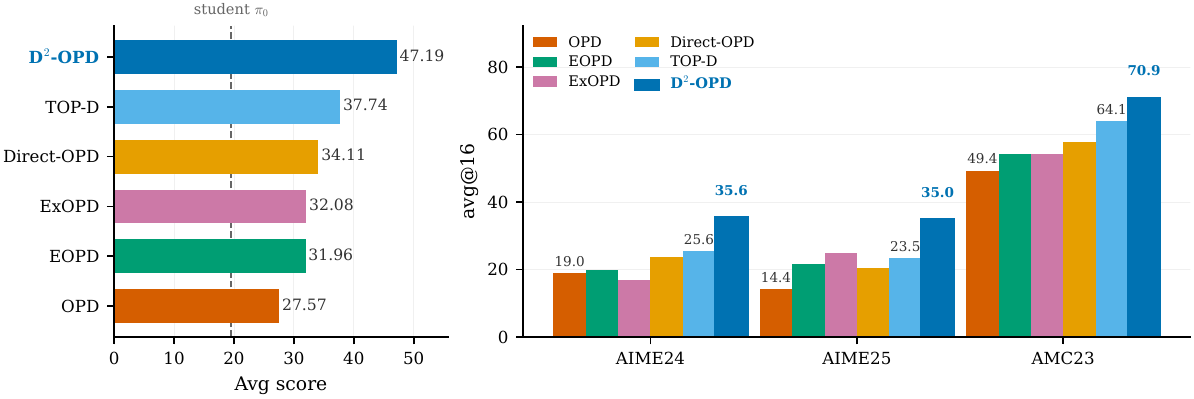}
\caption{
\textbf{Qwen3-8B \(\rightarrow\) Qwen3-4B-Base distillation.}
\textbf{Left:} average score across AIME24, AIME25, and AMC23 using avg@16 (mean accuracy over 16 sampled responses per problem), with the
initial student shown as a dashed reference. \ours{} reaches 47.19, 19.62 points higher than
standard OPD.
\textbf{Right:} per-benchmark results. The gain is consistent across all three benchmarks.
Numbers are from Table~\ref{tab:settingA}.}
\label{fig:headline}
\end{figure*}

\section{Introduction}
\label{sec:intro}

On-policy distillation (OPD) trains a student language model on its own responses: the student generates a response and, at each prefix it visits, learns from the teacher's next-token distribution \citep{gkd,minillm,opd}.
Training on student-generated prefixes reduces the mismatch between the responses seen in training and those the student generates at inference \citep{dagger,imitkd}.
Prior work has studied several aspects of on-policy distillation, including the efficient use of student-generated responses \citep{distillm,distillm2},
teacher entropy \citep{eopd}, teacher--student disagreement \citep{tropd},
and training stability \citep{topd}. These studies show that the way teacher supervision is constructed and
transferred matters beyond the choice of teacher alone.

However, the teacher's distribution may not be the best target: prior work shows that a more capable teacher does not necessarily yield a better student \citep{teacherassistant,rethinking}.
Extrapolative OPD aims to construct a better target beyond the teacher by amplifying the teacher's log-probability shift from a reference policy.
If the teacher was obtained from the reference by KL-regularized post-training, this shift equals its regularized \(Q\)-function up to scale and a prefix-dependent constant \citep{regularizedmdp,rtoq}, so amplifying it moves probability toward tokens with higher teacher \(Q\)-values.
Constructing an extrapolative target involves two decisions: the direction in which the target moves away from the teacher, and the distance it moves.
In existing methods, the distance is tied to the direction.
For example, ExOPD \citep{exopd} fixes the direction through its reference, but under a fixed coefficient its distance varies with the scale of the teacher--reference gap across prefixes; in Direct-OPD \citep{directopd}, changing the batch-shared adaptive coefficient generally changes both the direction and the distance.
In both cases, the distance follows from the coefficient and the log-ratios at each prefix, rather than from how far the target can be trusted beyond the teacher.
We show that, under fixed-coefficient extrapolation, the target can place substantial probability on tokens the teacher rarely predicts, even when the error in the extrapolation signal is small on average under the teacher.

To address this coupling, we propose \ours{}, which decouples distance from direction.
First, a per-prefix hard KL constraint bounds the target's distance from the teacher, independently of the scale of the extrapolation signal; this also bounds the probability the target can assign to tokens the teacher rarely predicts.
Second, we construct the direction by combining the teacher's post-training shift and the initial student--teacher gap.
The weights adapt at each prefix, giving more weight to the nearer signal.
Finally, forward KL transfers the target while maintaining non-vanishing gradients for tokens rarely generated by the student.

We evaluate \ours{} on three Qwen3 teacher--student pairs covering a setting where standard OPD
performs well, a setting for which prior work recommends a supervised cold start \citep{rethinking}, and a setting
without a separate teacher reference. \ours{} achieves the highest average score  across AIME24, AIME25, and AMC23 in all three settings. In Qwen3-8B \(\rightarrow\)
Qwen3-4B-Base distillation, it outperforms OPD by 19.62 points and the strongest baseline by 9.45
points. In the cold-start setting, it exceeds the released SFT + OPD recipe without supervised fine-tuning.
Ablations confirm the contributions of the distance constraint, the combined direction, and forward-KL distillation.
Response-length and entropy analyses indicate that the gains are not
merely due to longer answers or more concentrated student policies.

\section{Related work}
\label{sec:related}

\paragraph{Distillation objectives.}
Knowledge distillation trains students using teacher predictions, with applications to policy learning
and reasoning
\citep{hinton2015,kim2016,policydistill,distral,kickstarting,stepbystep,socratic}.
On-policy imitation and distillation reduce the mismatch between training and inference by training
on the learner's own trajectories \citep{dagger,imitkd,gkd}.
Related work studies divergence choices and response reuse for transferring teacher supervision
\citep{fdistill,minillm,distillm,distillm2}.

\paragraph{Teacher--student gaps.}
A larger or more capable teacher does not necessarily produce a better student
\citep{teacherassistant,atkd,taid,bornagain,weakstrong,skd}. 
EOPD \citep{eopd} adds forward KL when teacher entropy is high, while TrOPD \citep{tropd} applies
forward KL where teacher--student disagreement is large.
\citet{rethinking} link training failure to low initial overlap between
the teacher's and student's high-probability token sets.

\paragraph{Extrapolation and distance constraints.}
Under KL-regularized optimization, the log-ratio between the optimized policy and its reference encodes the regularized \(Q\)-function \citep{rlhf,regularizedmdp,dpo,rtoq}.
Extrapolative methods use this relation to construct new distillation targets.
ExOPD uses a fixed coefficient to scale a direction set by its reference \citep{exopd}.
Direct-OPD transfers the teacher's post-training shift through an objective anchored at the initial student, with an adaptive regularization coefficient applied uniformly across the batch \citep{directopd}.
Changing this coefficient generally changes both the target's direction and its distance relative to the teacher.
TOP-D constructs a proximal teacher and uses trust-region updates \citep{topd}; ERPD constrains
distillation relative to the rollout policy \citep{erpd}.

\section{Preliminaries}
\label{sec:prelim}

Let \(\mathcal D\) be the training prompt distribution and \(\pi_\theta\) the current student with parameters \(\theta\).
For a prompt \(x\sim\mathcal D\), we sample \(G\) responses \(y_1,\ldots,y_G\) from the student.
The index \(i\) identifies a response, \(t\) a token position, and \(|y_i|\) its response length.
Writing \(y_{i,<t}\) for the tokens before position \(t\), we define the prefix \(h_{i,t}=(x,y_{i,<t})\).
We use \(y\), \(h\), and \(a\) for a generic response, prefix, and candidate next token.
We denote the initial student by \(\pi_0\), the post-trained teacher by \(q^+\), and the checkpoint from which the teacher was post-trained by \(t_{\mathrm{ref}}\).
For any policy \(\pi\), \(\pi(a\mid h)\) denotes a token probability and \(\pi(\cdot\mid h)\) the next-token distribution, while \(\pi(y\mid x)\) denotes a response probability.
Let \(K\) be the number of candidate tokens and \(S(h)\) the top-\(K\) tokens under the teacher distribution \(q^+(\cdot\mid h)\).
For any token set \(B\), write \(\pi(B\mid h)=\sum_{a\in B}\pi(a\mid h)\).
The distribution renormalized on \(S(h)\) is \(\pi_S(a\mid h)=\pi(a\mid h)/\pi(S(h)\mid h)\) for \(a\in S(h)\), and zero otherwise.
In particular, \(q_S^+(\cdot\mid h)\) and \(\pi_{\theta,S}(\cdot\mid h)\) denote the renormalized teacher and student distributions, respectively, renormalized over the same top-\(K\) token set \(S(h)\).

\subsection{On-policy distillation}
\label{sec:prelim:opd}

Reverse-KL OPD \citep{minillm,opd} minimizes the loss \(\mathcal L_{\mathrm{OPD}}(\theta)\),
\begin{equation}
\label{eq:opd}
\mathcal L_{\mathrm{OPD}}(\theta)
=\mathbb E_{x\sim\mathcal D}
\left[D_{\mathrm{KL}}\!\left(\pi_\theta(\cdot\mid x)\,\|\,q^+(\cdot\mid x)\right)\right].
\end{equation}
At a student-generated prefix, the token reward is
\(r_\theta(h,a)=\log(q^+(a\mid h)/\pi_\theta(a\mid h))\).
Its expectation under the student is the negative reverse KL:
\[
\underset{a\sim\pi_\theta(\cdot\mid h)}{\mathbb E}[r_\theta(h,a)]
=-D_{\mathrm{KL}}(\pi_\theta(\cdot\mid h)\|q^+(\cdot\mid h)).
\]
Maximizing the expected sum of token
rewards is equivalent to minimizing Eq.~\eqref{eq:opd}.

\subsection{Extrapolative targets}
\label{sec:prelim:gopd}
\label{sec:prelim:qfunction}

For a reference \(b\), OPD is KL-regularized RL with reward
\(\log(q^+(y\mid x)/b(y\mid x))\) and regularizer anchored at \(b(\cdot\mid x)\). Scaling that reward by \(\lambda>0\)
gives the ExOPD target \(\pi^\star_\lambda\) \citep{exopd},
\begin{equation}
\label{eq:gopd-opt}
\pi^\star_\lambda(y\mid x)\propto
b(y\mid x)\left(\frac{q^+(y\mid x)}{b(y\mid x)}\right)^\lambda.
\end{equation}
It recovers the teacher at \(\lambda=1\) and extrapolates along
\(\log(q^+(y\mid x)/b(y\mid x))\) for \(\lambda>1\).

\begin{figure*}[!t]
\centering
\includegraphics[width=0.96\textwidth]{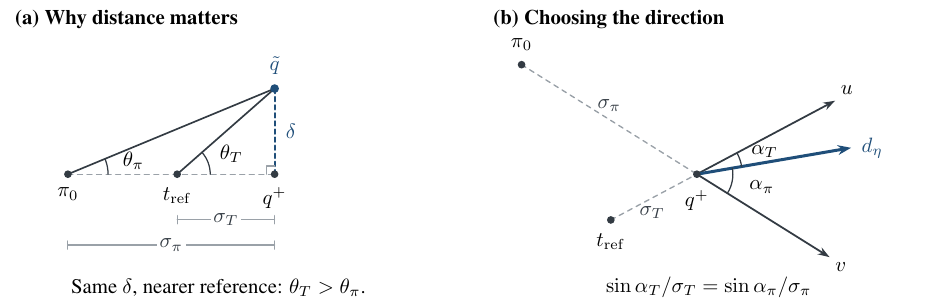}
\caption{\textbf{Distance-aware extrapolation direction.}
All quantities are shown at a fixed prefix \(h\), with conditioning omitted.
The distances from the teacher \(q^+\) to \(t_{\mathrm{ref}}\) and \(\pi_0\) are \(\sigma_T\) and \(\sigma_\pi\), with corresponding unit directions \(u\) and \(v\).
\textbf{(a) \(\theta_T>\theta_\pi\):} the same orthogonal displacement \(\delta\) from \(q^+\) to \(\tilde q\) produces a larger angular deviation at the nearer reference.
\textbf{(b)} With \(\sigma_T<\sigma_\pi\), the combined direction \(d_\eta\) lies closer to \(u\) than to \(v\), giving \(\alpha_T<\alpha_\pi\).}
\label{fig:direction}
\end{figure*}

The teacher reference gives this direction a policy-improvement interpretation.
Let \(\Delta_T(h,a)=\log(q^+(a\mid h)/t_{\mathrm{ref}}(a\mid h))\) be the teacher's post-training shift.
If \(q^+\) solves a KL-regularized objective relative to \(t_{\mathrm{ref}}\) with coefficient \(\beta_T>0\), its regularized \(Q\)-function \(Q^+(h,a)\) satisfies the following identity for an action-independent term \(c(h)\) \citep{regularizedmdp,rtoq}:
\begin{equation}
\label{eq:deltat}
\beta_T\Delta_T(h,a)
=Q^+(h,a)-c(h).
\end{equation}
\(\Delta_T(h,a)\) and \(Q^+(h,a)\) rank tokens identically.
The teacher's post-training shift can therefore serve as a signal for further policy improvement.
Direct-OPD uses \(\Delta_T(h,a)\) as its token reward and anchors its regularizer at \(\pi_0\), with regularization coefficient \(\alpha>0\).
Its optimal target \(\pi^\star\) is \citep{directopd}
\begin{equation}
\label{eq:dopd-opt}
\pi^\star(y\mid x)\propto
\pi_0(y\mid x)\left(\frac{q^+(y\mid x)}{t_{\mathrm{ref}}(y\mid x)}\right)^{1/\alpha}.
\end{equation}
This target need not lie on the teacher-reference ray in Eq.~\eqref{eq:gopd-opt}.
In both objectives, distance from \(q^+(\cdot\mid x)\) depends on the log-ratio scale as well as the
coefficient. Appendix~\ref{app:prior-objectives} gives the full objectives.

\section{Method}
\label{sec:method}

The objectives in Section~\ref{sec:prelim:gopd} tie two decisions together: the direction in which the
target leaves the teacher, and how far it travels. We construct the extrapolation direction and separately bound the target's distance from the teacher at each prefix.

For a prefix \(h\) visited by a student rollout, we build a target distribution \(\tilde q(\cdot\mid h)\) along a direction derived
from two reference signals, at a distance we bound explicitly, and train the student toward it with
forward KL, using the loss \(\mathcal L(\theta)\):
\begin{equation}
\label{eq:outer-loss}
\resizebox{\columnwidth}{!}{$\displaystyle
\mathcal L(\theta)
=\underset{\substack{x\sim\mathcal D\\
\{y_i\}_{i=1}^{G}\sim\pi_\theta(\cdot\mid x)}}{\mathbb E}
\Biggl[\frac{1}{\sum_{i=1}^{G}|y_i|}
\sum_{i=1}^{G}\sum_{t=1}^{|y_i|}
D_{\mathrm{KL}}\!\left(
\tilde q(\cdot\mid h_{i,t})\,\|\,\pi_\theta(\cdot\mid h_{i,t})
\right)\Biggr].$}
\end{equation}
We next explain the rationales behind each of our design choices.
Section~\ref{sec:trust} bounds how far \(\tilde q(\cdot\mid h)\) may move from the teacher, through a hard radius
\(\epsilon\ge0\) and a soft coefficient \(0<\kappa\le1\). Section~\ref{sec:direction} determines the combined direction
\(d_\eta(h,a)\), where \(\eta(h)\in[0,1]\) is derived to weight the student signal by minimizing relative deviation from both signals. Section~\ref{sec:forwardkl} explains why the divergence in
Eq.~\eqref{eq:outer-loss} is forward rather than reverse KL.

\subsection{Limiting the extrapolation distance}
\label{sec:trust}

As discussed in Section~\ref{sec:prelim:qfunction}, existing extrapolation coefficients control the distance from \(q^+(\cdot\mid h)\) only indirectly: a coefficient multiplies a log-ratio whose scale varies across prefixes, so one value can hold the target close to \(q^+(\cdot\mid h)\) at one prefix and carry it much farther at another.
The distance is worth controlling on its own, because the teacher is a policy we can observe and evaluate, whereas a target farther from \(q^+(\cdot\mid h)\) rests more heavily on extrapolation.
For a candidate target distribution \(q(\cdot\mid h)\) supported on \(S(h)\), let \(\mathcal K(q;h)=D_{\mathrm{KL}}(q(\cdot\mid h)\|q_S^+(\cdot\mid h))\). We impose \(\mathcal K(q;h)\le\epsilon\) at every prefix, which bounds how far the target may move. 
To express token preference in the target distribution, let \(d(h,a)\) be a per-token score function on \(S(h)\), whose construction is introduced in Section~\ref{sec:direction}.
We choose \(\tilde q(\cdot\mid h)\) to increase the expected value of \(d(h,a)\) while penalizing its KL distance from the renormalized teacher \(q_S^+(\cdot\mid h)\).
The coefficient \(\kappa\) controls this trade-off, and the constraint limits the distance to \(\epsilon\).
Therefore, the target distribution \(\tilde q(\cdot\mid h)\) is defined as:
\begin{equation}
\label{eq:target}
\begin{aligned}
\tilde q(\cdot\mid h) = \argmax_{q(\cdot\mid h)}\quad &
 (1-\kappa)\underset{a\sim q(\cdot\mid h)}{\mathbb E}[d(h,a)]\\
& -\kappa\mathcal K(q;h)\\
\text{subject to}\quad &\mathcal K(q;h)\le\epsilon.
\end{aligned}
\end{equation}

\paragraph{Solving the constrained problem.}

Let \(\lambda(h)\ge0\) be the Lagrange multiplier on the KL constraint and \(\beta(h)\) the extrapolation coefficient.
The optimum of Eq.~\eqref{eq:target} has the form
\begin{equation}
\label{eq:constrained-target-solver}
\begin{aligned}
\tilde q(a\mid h)&\propto q_S^+(a\mid h)\exp\!\left[\beta(h) d(h,a)\right],\\
\beta(h)&=\frac{1-\kappa}{\kappa+\lambda(h)}.
\end{aligned}
\end{equation}
The target is normalized over \(S(h)\) and zero elsewhere.
The unconstrained solution yields \(\lambda(h)=0\) and coefficient \(\beta_{\max}=(1-\kappa)/\kappa\).
At \(\kappa=1\), the target reduces to \(q_S^+(\cdot\mid h)\) while smaller values allow stronger extrapolation.

For \(0<\kappa<1\), we retain the unconstrained target if it satisfies the bound; otherwise, we perform 24 bisection iterations on \([0,\beta_{\max}]\), where \(D_{\mathrm{KL}}(\tilde q(\cdot\mid h)\|q_S^+(\cdot\mid h))\) is nondecreasing in \(\beta(h)\), to find the constraint boundary.
We process prefixes in parallel on the teacher's top-\(K\) support.
The search is computationally inexpensive.\footnote{On a single A100 40GB GPU, the search takes about 7.2~ms for 32,768 prefixes, equivalent to eight responses of 4,096 tokens each, with \(K=16\) in FP32. Including direction construction takes about 9.3~ms.}

ExOPD observes longer responses under stronger extrapolation and attributes this partly to potential length bias in the implicit reward~\citep{exopd}.
We propose another possible mechanism: extrapolation may amplify overestimated scores for low-probability teacher tokens, potentially encouraging repetitive or prolonged reasoning. 
The KL bound limits the probability mass assigned to such tokens.

\begin{theoryproposition}
\label{prop:theory:rare-errors}
At a fixed prefix \(h\), let \(B\subseteq S(h)\) be a set of tokens with renormalized teacher probability \(\rho(h)=q_S^+(B\mid h)\), where \(0<\rho(h)<1\).
Every target \(q(\cdot\mid h)\) satisfying \(\mathcal K(q;h)\le\epsilon\) obeys
\begin{equation}
\label{eq:theory:rare-bound}
q(B\mid h)\le\frac{\epsilon+\log2}{\log(1/\rho(h))}.
\end{equation}
\end{theoryproposition}
For fixed \(\epsilon\), this bound tends to zero as the teacher probability \(\rho(h)\) tends to zero.
Appendix~\ref{app:theory} contains the proof and expands on the connection to response-length growth.
It also shows that fixed-coefficient extrapolation can assign substantial probability to these tokens even when the score error is small in teacher-weighted mean square.

\subsection{Determining the extrapolation direction}
\label{sec:direction}

The teacher's post-training shift and the initial student--teacher gap typically point in different directions.
Although both provide useful guidance, constructing a target requires combining them into a single direction.
As shown in Section~\ref{sec:prelim:qfunction}, Direct-OPD's regularization coefficient \(\alpha\) controls both direction and distance, without a separate criterion for combining the signals.
We therefore introduce an explicit criterion that balances the direction's relative deviation from the two signals.

We use the teacher-reference signal \(\Delta_T(h,a)=\log(q^+(a\mid h)/t_{\mathrm{ref}}(a\mid h))\) and the student signal \(\Delta_{\pi_0}(h,a)=\log(q^+(a\mid h)/\pi_0(a\mid h))\).
For convenience, we center both signals so that their variances equal their squared norms under \(q_S^+(\cdot\mid h)\).
At each prefix, for \(j\in\{T,\pi_0\}\), we define \(\Delta_j^c(h,a)=\Delta_j(h,a)-\mathbb E_{b\sim q_S^+(\cdot\mid h)}[\Delta_j(h,b)]\).

Let \(\sigma_T^2(h)\) and \(\sigma_\pi^2(h)\) denote the variances of the teacher-reference and student signals, respectively, under \(q_S^+(\cdot\mid h)\).
In centered log-probability space weighted by \(q_S^+(\cdot\mid h)\), each signal's variance is the squared distance from the teacher to its reference.
We refer to \(\sigma_T(h)\) and \(\sigma_\pi(h)\) as reference distances and normalize each squared deviation by its corresponding squared reference distance.
When both reference distances are positive, let
\(\mathcal R_T(d;h)=\mathbb E_{a\sim q_S^+(\cdot\mid h)}[(d(h,a)-\Delta_T^c(h,a))^2]/\sigma_T^2(h)\)
and define \(\mathcal R_\pi(d;h)\) analogously using \(\Delta_{\pi_0}^c(h,a)\) and \(\sigma_\pi^2(h)\).
We define \(d_\eta(h,\cdot)\) as:
\begin{equation}
\label{eq:relative-direction}
\begin{aligned}
d_\eta(h,\cdot) = \argmin_{d(h,\cdot)}\quad &\mathcal R_T(d;h)+\mathcal R_\pi(d;h)\\
\text{subject to}\quad &\mathbb E_{a\sim q_S^+(\cdot\mid h)}[d(h,a)]=0.
\end{aligned}
\end{equation}
\begin{samepage}
\begin{theoryproposition}
\label{prop:theory:relative-direction}
When both reference distances \(\sigma_T(h)\) and \(\sigma_\pi(h)\) are positive, the objective above has the unique minimizer
\begin{align}
d_\eta(h,a)&=(1-\eta(h))\Delta_T^c(h,a)\!+\!\eta(h)\Delta_{\pi_0}^c(h,a),\notag\\
\eta(h)&=\frac{\sigma_T^2(h)}{\sigma_T^2(h)+\sigma_\pi^2(h)}.
\label{eq:direction-implementation}
\end{align}
\end{theoryproposition}
\end{samepage}
The proof is given in Appendix~\ref{app:theory}.
The weight \(\eta(h)\) is determined by the two signals at each prefix, giving more weight to the nearer reference.

For the geometric interpretation, let \(u(h,a)=\Delta_T^c(h,a)/\sigma_T(h)\) and \(v(h,a)=\Delta_{\pi_0}^c(h,a)/\sigma_\pi(h)\) be the unit directions.
The solution satisfies
\begin{equation}
\label{eq:weighted-bisector}
d_\eta(h,a)\propto\sigma_\pi(h)u(h,a)+\sigma_T(h)v(h,a).
\end{equation}
Figure~\ref{fig:direction} illustrates how this combination places the direction closer to that of the nearer reference.

In our implementation, we compute the closed-form solution directly without forming \(u(h,a)\) and \(v(h,a)\), avoiding division by either reference distance.
We use Eq.~\eqref{eq:direction-implementation} when both reference distances exceed \(10^{-3}\).
If only one is at most \(10^{-3}\), we omit that signal and use the other centered signal.
If both are at most \(10^{-3}\), we set \(d_\eta(h,a)=0\), so the target remains \(q_S^+(\cdot\mid h)\).

We set \(d(h,a)=d_\eta(h,a)\) in Eq.~\eqref{eq:target}.
Following ExOPD, we can equivalently express this direction using a constructed reference \(t_\eta(\cdot\mid h)\) on \(S(h)\):
\begin{equation}
\label{eq:teta}
t_\eta(a\mid h)
\propto
q_S^+(a\mid h)\exp[-d_\eta(h,a)].
\end{equation}
After normalization over \(S(h)\), \(\log(q_S^+(a\mid h)/t_\eta(a\mid h))\) equals \(d_\eta(h,a)\) up to a token-independent constant.
This constructed reference requires no additional model.

\subsection{Transferring the target to the student}
\label{sec:forwardkl}

The constructed target can place probability to tokens that are rare under the student. We use forward
KL because its gradient remains non-vanishing in this case. The forward-KL gradient with respect to the student logit is \(\pi_\theta(a\mid h)-\tilde q(a\mid h)\).
For tokens with positive target probability, it approaches \(-\tilde q(a\mid h)\) as \(\pi_\theta(a\mid h)\to0\), whereas the reverse-KL gradient vanishes.

In practice, \(S(h)\) contains the top-\(K\) tokens under the teacher distribution \(q^+(\cdot\mid h)\). We construct and renormalize \(\tilde q(\cdot\mid h)\) on \(S(h)\), but keep the student distribution over the full vocabulary.
Estimating forward KL from student-sampled tokens can have high variance.
The importance weights \(\tilde q(a\mid h)/\pi_\theta(a\mid h)\) can be large when the target assigns substantial probability to tokens the student rarely samples.
We therefore compute the exact sum over all tokens in \(S(h)\) at each visited prefix.
With support mass \(\pi_\theta(S(h)\mid h)\) and renormalized student \(\pi_{\theta,S}(\cdot\mid h)\) defined in Section~\ref{sec:prelim}, we have
\begin{equation}
\label{eq:fkl-support}
\resizebox{\dimexpr\columnwidth-3em\relax}{!}{$\displaystyle
\begin{aligned}
&D_{\mathrm{KL}}(\tilde q(\cdot\mid h)\|\pi_\theta(\cdot\mid h))=\\
&-\log \pi_\theta(S(h)\mid h)
+D_{\mathrm{KL}}(\tilde q(\cdot\mid h)\|\pi_{\theta,S}(\cdot\mid h)).
\end{aligned}$}
\end{equation}
The first term shifts student probability mass onto \(S(h)\), while the second matches its relative
distribution within it. Renormalizing the student removes the first term and its mass-transfer effect.
Unlike Eq.~\eqref{eq:fkl-support}, reverse KL has no such mass-transfer term and only adjusts relative probabilities within \(S(h)\). We examine the importance of this forward-KL choice in the ablation experiments in Section~\ref{sec:experiments}.

\section{Experiments}
\label{sec:experiments}
\begin{table}[!b]
\centering
\small
\setlength{\tabcolsep}{2pt}
\renewcommand{\arraystretch}{1.15}
\caption{\textbf{Setting A}: Qwen3-8B \(\rightarrow\) Qwen3-4B-Base.}
\label{tab:settingA}
\begin{tabularx}{\columnwidth}{@{}lRRRRr@{}}
\arrayrulecolor{tableNavy}\toprule[0.9pt]\arrayrulecolor{tableRule}
\rowcolor{white}
\textcolor{black}{\textbf{Method}} & \textcolor{black}{\shortstack{\textbf{AIME}\\\textbf{24}}} & \textcolor{black}{\shortstack{\textbf{AIME}\\\textbf{25}}} & \textcolor{black}{\shortstack{\textbf{AMC}\\\textbf{23}}} & \textcolor{black}{\textbf{Avg}} & \textcolor{black}{\(\boldsymbol{\Delta}\)} \\
\midrule
\rowcolor{refband}
Student \(\pi_0\) & 12.50 &  6.25 & 40.00 & \cellcolor{summaryband}19.58 & \(-7.99\) \\
\rowcolor{refband}
Teacher \(q^+\)   & 75.00 & 64.17 & 94.22 & \cellcolor{summaryband}77.80 & \(+50.23\) \\
\midrule
OPD    & 18.96 & 14.38 & 49.38 & \cellcolor{summaryband}27.57 & --- \\
EOPD  & 20.00 & 21.67 & 54.22 & \cellcolor{summaryband}31.96 & \(+4.39\) \\
ExOPD  & 17.08 & \underline{24.79} & 54.37 & \cellcolor{summaryband}32.08 & \(+4.51\) \\
Direct-OPD  & 23.75 & 20.62 & 57.97 & \cellcolor{summaryband}34.11 & \(+6.54\) \\
TOP-D  & \underline{25.62} & 23.54 & \underline{64.06} & \cellcolor{summaryband}\underline{37.74} & \(+10.17\) \\
\rowcolor{oursband}
\textcolor{oursink}{\textbf{\ours{}}}           & \best{35.62} & \best{35.00} & \best{70.94} & \cellcolor{ourssummary}\best{47.19} & \gain{+19.62} \\
\bottomrule
\end{tabularx}\arrayrulecolor{black}
\end{table}

\subsection{Experimental setup}
\label{sec:setup}

\paragraph{Models.}
We evaluate three teacher--student pairs covering three complementary settings. All models are from the
Qwen3 family \citep{qwen3}.
(1) Setting~A distills Qwen3-8B into Qwen3-4B-Base, with
\(t_{\mathrm{ref}}=\) Qwen3-8B-Base. Under this setting, standard OPD performs well and a genuine
teacher reference is available.
(2) Setting~B distills Qwen3-4B (non-thinking) into Qwen3-1.7B-Base, with
\(t_{\mathrm{ref}}=\) Qwen3-4B-Base. \citet{rethinking} report that OPD leads to poor performance under this setting and recommend a supervised finetuning as a cold start, while a genuine teacher reference is available.
(3) Setting~C distills Qwen3-30B-A3B-Instruct-2507 into Qwen3-8B-Base. Standard OPD performs well in
this setting, but no teacher reference checkpoint is available, so we set
\(t_{\mathrm{ref}}=\pi_0\). There is then only one extrapolative direction, leaving \(\epsilon\) and
\(\kappa\) as the active controls.

\paragraph{Training data.}
Settings~A and C train on the full DAPO-Math-17K dataset \citep{dapo}. Setting~B trains on approximately 30K prompts from the math-domain subset of OpenThoughts3-1.2M \citep{openthoughts}. Every run trains for one epoch.

\paragraph{Evaluation.}
We evaluate on AIME24, AIME25, and AMC23 (\citeauthor{maa_aime}, \citeyear{maa_amc}, \citeyear{maa_aime}), reporting avg@16 with a shared maximum evaluation response length of 31{,}744 tokens.
We report results from the final checkpoint of one training run per configuration.

\paragraph{Baselines.}
We compare against OPD \citep{opd}, EOPD \citep{eopd}, TOP-D \citep{topd}, ExOPD \citep{exopd}, and
Direct-OPD \citep{directopd}.
We use the on-policy variant of TOP-D, the reward-correction variant of ExOPD, and the published adaptive \(\alpha\) controller of Direct-OPD.
For method-specific hyperparameters, we use the authors' recommended values.
Appendix~\ref{app:baseline-search} gives implementation details and parameter-search results.

\paragraph{Hyperparameters.}
All methods share the default training setup of \citet{rethinking}, except for a maximum training response length of 4{,}096 tokens.
Appendix~\ref{app:hyperparams} provides training and evaluation details.

\paragraph{Method parameters.}
Unless otherwise stated, we use \(\kappa=0.05\), with \(\epsilon=1\) in Settings~A and B and \(\epsilon=4\) in Setting~C.
Appendix~\ref{app:sweeps} describes parameter selection and sensitivity.

\begin{figure*}[!t]
\centering
\hspace*{0.05\textwidth}%
\makebox[0.316\textwidth]{\small Setting A}%
\makebox[0.316\textwidth]{\small Setting B}%
\makebox[0.316\textwidth]{\small Setting C}\par
\includegraphics[width=\textwidth,trim=0 0 0 217bp,clip]{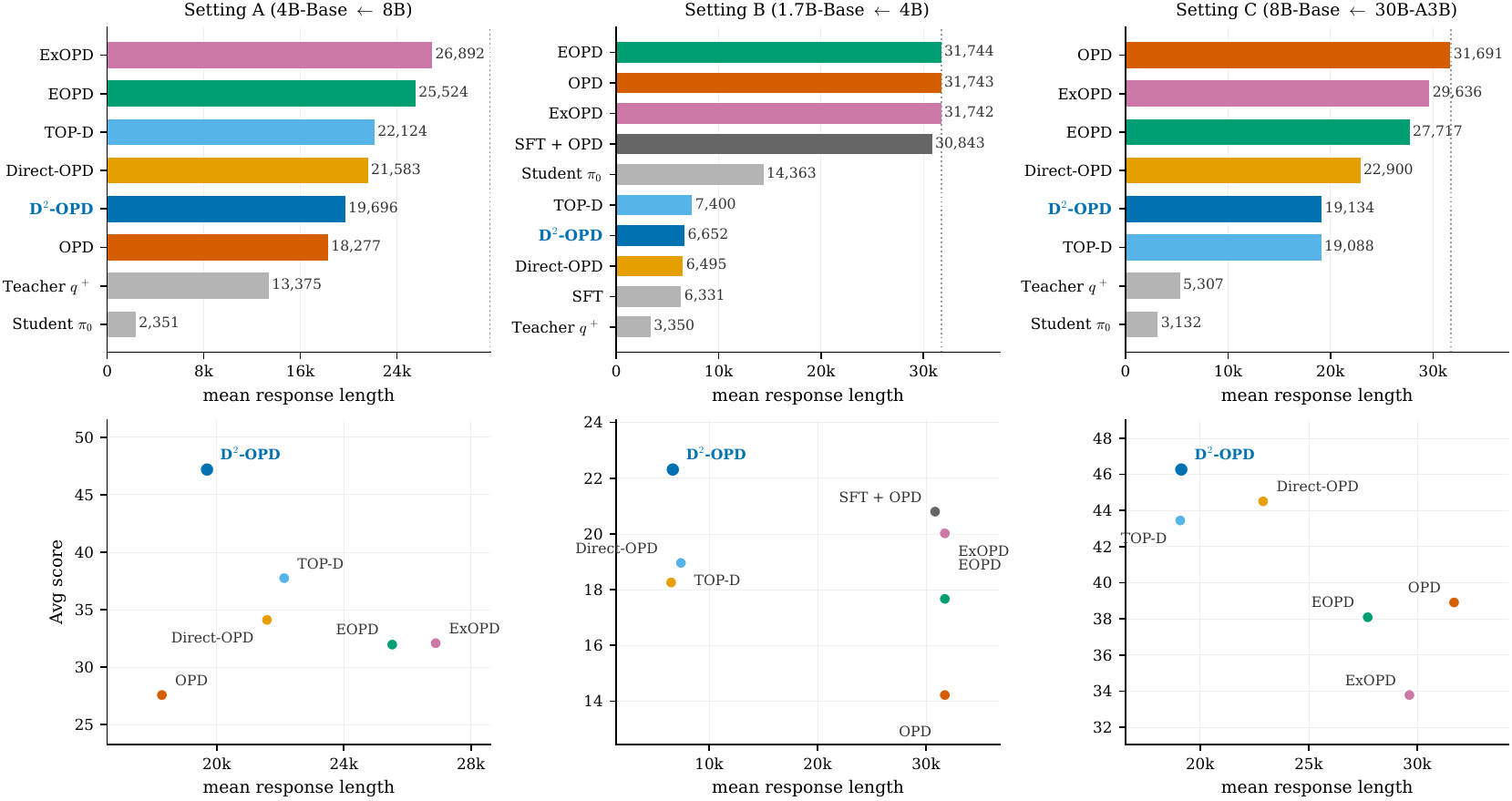}
\caption{\textbf{Longer responses do not explain the gain.}
Average score versus mean response length for the trained methods across the three settings.
Appendix~\ref{app:response-length} includes the full response-length comparison.}
\label{fig:response-length-full}
\end{figure*}

\subsection{Main results}
\label{sec:main}

Tables~\ref{tab:settingA}--\ref{tab:settingC} report the main results across the three settings.
In these three tables, \textbf{Avg} is the mean score across the three benchmarks, and \(\Delta\) denotes its difference from OPD. Bold and underline indicate
the best and second-best results, respectively, among trained methods.

\paragraph{Setting A.}
OPD improves the initial student's average score from 19.58 to 27.57.
Our method reaches 47.19, improving OPD by 19.62 points and the strongest baseline,
TOP-D, by 9.45 points. It leads on all three benchmarks.

\paragraph{Setting B.}
We also compare the SFT checkpoint and SFT + OPD recipe released by \citet{rethinking}.

\begin{table}[tbp]
\centering
\small
\setlength{\tabcolsep}{2pt}
\renewcommand{\arraystretch}{1.15}
\caption{\textbf{Setting B}: Qwen3-4B \(\rightarrow\) Qwen3-1.7B-Base.}
\label{tab:settingB}
\begin{tabularx}{\columnwidth}{@{}lRRRRr@{}}
\arrayrulecolor{tableNavy}\toprule[0.9pt]\arrayrulecolor{tableRule}
\rowcolor{white}
\textcolor{black}{\textbf{Method}} & \textcolor{black}{\shortstack{\textbf{AIME}\\\textbf{24}}} & \textcolor{black}{\shortstack{\textbf{AIME}\\\textbf{25}}} & \textcolor{black}{\shortstack{\textbf{AMC}\\\textbf{23}}} & \textcolor{black}{\textbf{Avg}} & \textcolor{black}{\(\boldsymbol{\Delta}\)} \\
\midrule
\rowcolor{refband}
Student \(\pi_0\)                & 0.83 &  1.04 & 12.66 & \cellcolor{summaryband}4.84 & \(-9.38\) \\
\rowcolor{refband}
Teacher \(q^+\)                  & 24.17 & 15.62 & 64.22 & \cellcolor{summaryband}34.67 & \(+20.45\) \\
\midrule
OPD                   &  6.25 &  5.62 & 30.78 & \cellcolor{summaryband}14.22 & ---        \\
SFT            &  7.71 &  4.58 & 40.00 & \cellcolor{summaryband}17.43 & \(+3.21\)  \\
SFT + OPD      & \best{14.58} & \best{10.00} & 37.81 & \cellcolor{summaryband}\underline{20.80} & \(+6.58\) \\
\midrule
EOPD                 &  4.79 & \underline{9.79} & 38.44 & \cellcolor{summaryband}17.67 & \(+3.45\) \\
Direct-OPD      &  8.12 &  6.67 & 40.00 & \cellcolor{summaryband}18.26 & \(+4.04\) \\
\shortstack[l]{OPD w.\\forward KL} &  8.54 &  6.46 & \underline{41.56} & \cellcolor{summaryband}18.85 & \(+4.63\) \\
TOP-D                &  8.54 &  7.08 & 41.25 & \cellcolor{summaryband}18.96 & \(+4.74\) \\
ExOPD               & \underline{13.33} &  6.88 & 39.84 & \cellcolor{summaryband}20.02 & \(+5.80\) \\
\rowcolor{oursband}
\textcolor{oursink}{\textbf{\ours{}}}                          &  9.79 &  7.29 & \best{49.84} & \cellcolor{ourssummary}\best{22.31} & \gain{+8.09} \\
\bottomrule
\end{tabularx}\arrayrulecolor{black}
\end{table}

\emph{OPD w. forward KL} uses Section~\ref{sec:forwardkl} without extrapolation or direction selection.
Forward KL alone raises OPD's average score from 14.22 to 18.85.
The full method reaches 22.31, exceeding ExOPD by 2.29 points and SFT + OPD by 1.51 points,
without a supervised cold start. SFT + OPD remains stronger on AIME24 and AIME25;
our method leads on AMC23.

\begin{table}[tbp]
\centering
\small
\setlength{\tabcolsep}{2pt}
\renewcommand{\arraystretch}{1.15}
\caption{\textbf{Setting C}: Qwen3-30B-A3B-Instruct-2507 \(\rightarrow\) Qwen3-8B-Base. Here \(t_{\mathrm{ref}}=\pi_0\), so the reference directions coincide.}
\label{tab:settingC}
\begin{tabularx}{\columnwidth}{@{}lRRRRr@{}}
\arrayrulecolor{tableNavy}\toprule[0.9pt]\arrayrulecolor{tableRule}
\rowcolor{white}
\textcolor{black}{\textbf{Method}} & \textcolor{black}{\shortstack{\textbf{AIME}\\\textbf{24}}} & \textcolor{black}{\shortstack{\textbf{AIME}\\\textbf{25}}} & \textcolor{black}{\shortstack{\textbf{AMC}\\\textbf{23}}} & \textcolor{black}{\textbf{Avg}} & \textcolor{black}{\(\boldsymbol{\Delta}\)} \\
\midrule
\rowcolor{refband}
Student \(\pi_0\) & 19.76 & 11.67 & 50.62 & \cellcolor{summaryband}27.35 & \(-11.56\) \\
\rowcolor{refband}
Teacher \(q^+\)   & 73.33 & 62.29 & 99.69 & \cellcolor{summaryband}78.44 & \(+39.53\) \\
\midrule
ExOPD  & 28.33 & 25.21 & 47.81 & \cellcolor{summaryband}33.78 & \(-5.13\) \\
EOPD  & 33.54 & 19.79 & 60.94 & \cellcolor{summaryband}38.09 & \(-0.82\) \\
OPD    & 30.00 & 25.00 & 61.72 & \cellcolor{summaryband}38.91 & --- \\
TOP-D  & 29.17 & \best{30.42} & 70.78 & \cellcolor{summaryband}43.45 & \(+4.54\) \\
Direct-OPD  & \best{35.42} & 25.62 & \underline{72.50} & \cellcolor{summaryband}\underline{44.51} & \(+5.60\) \\
\rowcolor{oursband}
\textcolor{oursink}{\textbf{\ours{}}}           & \best{35.42} & \underline{27.92} & \best{75.47} & \cellcolor{ourssummary}\best{46.27} & \gain{+7.36} \\
\bottomrule
\end{tabularx}\arrayrulecolor{black}
\end{table}

\paragraph{Setting C.}
We set \(t_{\mathrm{ref}}=\pi_0\) in ExOPD, Direct-OPD, and our method. Their reference
directions coincide, so this setting isolates control along a fixed direction.
The constraint binds more often here; we use \(\epsilon=4\) (Appendix~\ref{app:sweeps}).
Our method reaches 46.27, improving OPD by 7.36 points and the strongest baseline,
Direct-OPD, by 1.76 points. TOP-D, which constructs an intermediate teacher from the
current student, also remains competitive.

Our method has the highest average score in all three settings, while the strongest
competing baseline changes across settings.

\subsection{Analysis}
\label{sec:analysis}

\begin{figure*}[!t]
\centering
\includegraphics[width=\textwidth,trim=0 410bp 0 0,clip]{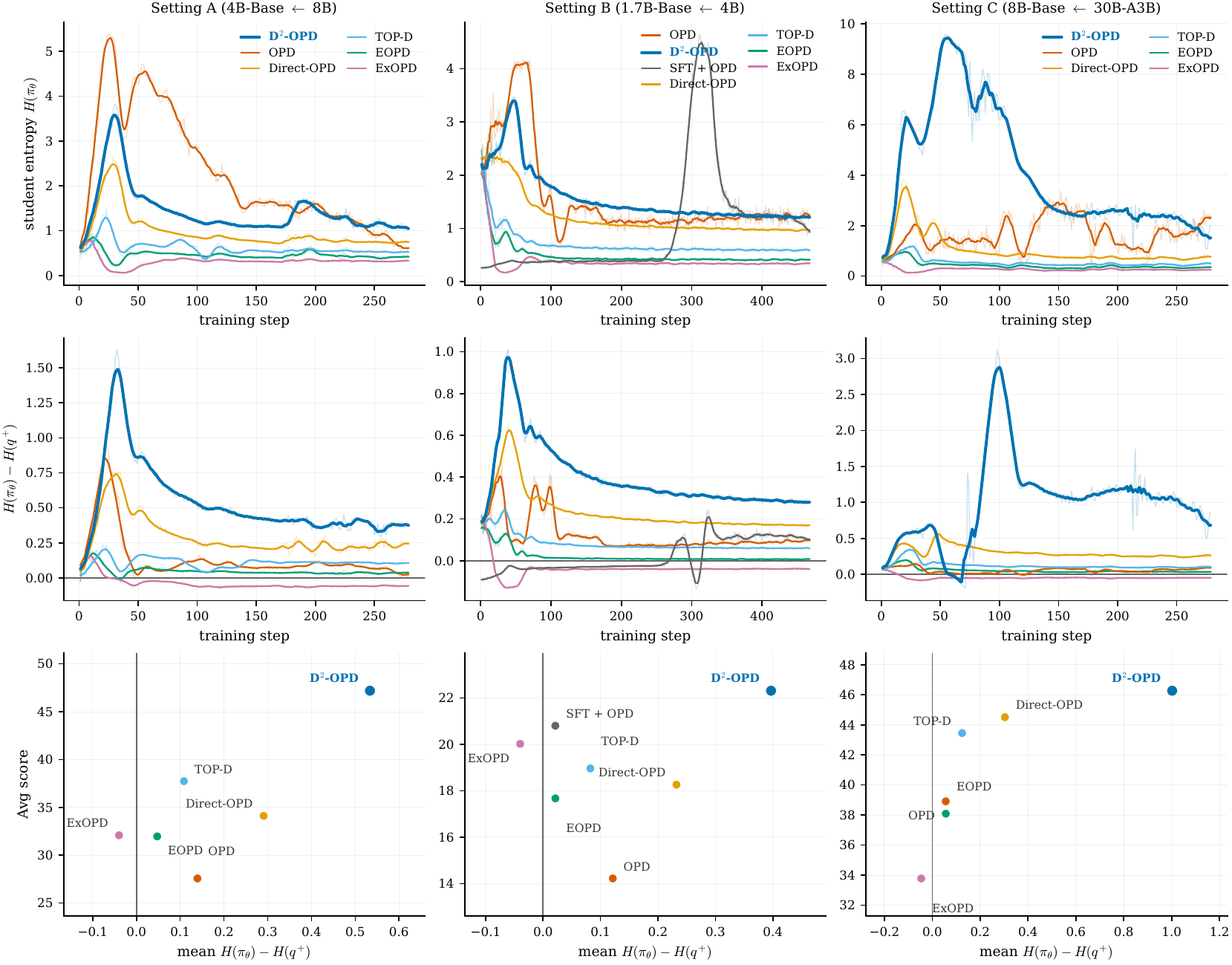}
\caption{\textbf{The gains are not due to lower entropy.} We measure full-vocabulary student entropy on each method’s on-policy prefixes.
Faint lines show per-step values; solid lines show a 9-step rolling mean.}
\label{fig:entropy-full}
\end{figure*}

\begin{figure}[!htb]
\centering
\includegraphics[width=\columnwidth]{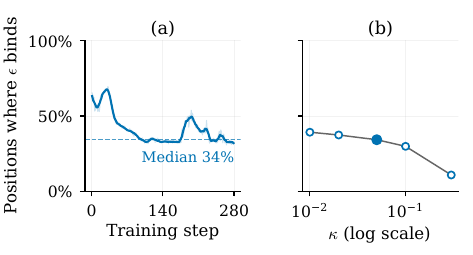}
\caption{\textbf{The \(\epsilon\)-constraint remains active.}
Fraction of token positions where the constraint binds in Setting~A, with \(\epsilon=1\).
\textbf{(a)} Training at \(\kappa=0.05\): per-step values (faint), a 9-step mean (solid),
and the second-half median (dashed).
\textbf{(b)} Second-half median versus \(\kappa\); the filled marker denotes \(\kappa=0.05\).}
\label{fig:constraint-active}
\end{figure}

\paragraph{How restrictive is \(\epsilon\)?}
To start with, we give a simple interpretation of the scale of \(\epsilon\). For a one-hot target \(\delta_a(\cdot\mid h)\) that assigns all probability to a token \(a\in S(h)\), \(D_{\mathrm{KL}}(\delta_a(\cdot\mid h)\|q_S^+(\cdot\mid h))=-\log q_S^+(a\mid h)\), which means that \(\epsilon=1\) permits the target to concentrate entirely on a token with \(q_S^+(a\mid h)\ge e^{-1}\approx0.37\). Although this makes \(\epsilon=1\) fairly permissive, Figure~\ref{fig:constraint-active} shows that it still binds a meaningful fraction of positions in practice. The constraint binds on more than 60\% of positions early in training and has a median active fraction of 34\% over the second half of training.
As expected, increasing \(\kappa\) makes the constraint less active, since the soft objective itself proposes less aggressive extrapolation. Even at \(\kappa=0.3\), the hard constraint still binds on about 11\% of positions.

\paragraph{Are the gains explained by longer responses?}
ExOPD observes that extrapolation can increase response length \citep{exopd}. We observe the same behavior for ExOPD, but not for extrapolative methods in general. In Figure~\ref{fig:response-length-full}, \ours{} and Direct-OPD
remain among the shorter methods across all three settings. TOP-D has a similar length and also remains well below the generation limit. In contrast, ExOPD and EOPD generally produce
much longer responses. It is the most obvious in Settings~B and C, where several methods nearly exhaust the 31{,}744-token budget while \ours{} does not.\footnote{Our evaluation length limit matches the limit used by \citet{rethinking,directopd} and is larger than those used by \citet{eopd,exopd,topd}.}
\ours{} achieves the highest average score in all three settings while generating shorter responses than several baselines.

\paragraph{Are the gains explained by lower entropy?}
A common intuition is that a lower-entropy policy is more concentrated and may therefore achieve higher single-response accuracy (pass@1). Figure~\ref{fig:entropy-full} shows that this does not explain our results.
Across all three settings, \ours{} remains among the higher-entropy students while achieving the best average score. TOP-D and Direct-OPD, which are also among the stronger baselines, generally
lie in the higher-entropy group as well. 
Appendix~\ref{app:entropy-gap} examines the student--teacher entropy gap.

\paragraph{Computational cost.}
Table~\ref{tab:computational-cost} reports time and throughput per training step excluding rollout generation.
\ours{} has roughly comparable training cost to OPD, taking 15\% more time per step.
Appendix~\ref{app:computational-cost} describes this experiment's setup.

\begin{table}[!htb]
\centering
\small
\setlength{\tabcolsep}{2pt}
\renewcommand{\arraystretch}{0.95}
\caption{Computational cost.}
\label{tab:computational-cost}
\begin{tabularx}{\columnwidth}{@{}lRRR@{}}
\arrayrulecolor{tableNavy}\toprule[0.9pt]\arrayrulecolor{tableRule}
\rowcolor{white}
\textcolor{black}{\textbf{Method}} & \textcolor{black}{\shortstack{\textbf{Time/step}\\(s)}} & \textcolor{black}{\shortstack{\textbf{Throughput}\\(tokens/s)}} & \textcolor{black}{\shortstack{\textbf{Relative}\\\textbf{time}}} \\
\midrule
OPD & 33.42 & 1,840 & \(1.00\times\) \\
Direct-OPD & 43.26 & 1,422 & \(1.29\times\) \\
TOP-D & 32.96 & 1,866 & \(0.99\times\) \\
\rowcolor{oursband}
\textcolor{oursink}{\textbf{\ours{}}} & 38.48 & 1,598 & \(1.15\times\) \\
\bottomrule
\end{tabularx}\arrayrulecolor{black}
\end{table}

\subsection{Ablations}
\label{sec:ablation}

\begin{table}[!htb]
\centering
\small
\setlength{\tabcolsep}{3pt}
\renewcommand{\arraystretch}{0.95}
\caption{Ablations in Setting~A. \textbf{Avg} is the mean over AIME24, AIME25, and AMC23; \(\Delta\) is relative to the full method.
Per-benchmark scores appear in Appendix~\ref{app:ablations}.}
\label{tab:ablation}
\begin{tabularx}{\columnwidth}{@{}llRR@{}}
\arrayrulecolor{tableNavy}\toprule[0.9pt]\arrayrulecolor{tableRule}
\rowcolor{white}
\textcolor{black}{\textbf{Component}} & \textcolor{black}{\textbf{Variant}} & \textcolor{black}{\textbf{Avg}} & \textcolor{black}{\(\boldsymbol{\Delta}\)} \\
\midrule
\rowcolor{oursband}
\multicolumn{2}{l}{\cellcolor{oursband}full} & \cellcolor{ourssummary}\best{47.19} & --- \\
\midrule
\multirow[t]{3}{*}{\textcolor{tableMuted}{constraint}}
  & \(- \kappa\)            & \cellcolor{summaryband}45.59 & \textcolor{tableNegative}{\(-1.60\)} \\
  & \(-\epsilon\)                         & \cellcolor{summaryband}43.30 & \textcolor{tableNegative}{\(-3.89\)} \\
  & \(-\)both                            & \cellcolor{summaryband}42.73 & \textcolor{tableNegative}{\(-4.46\)} \\
\addlinespace[3pt]\cmidrule(l){2-4}
\multirow[t]{2}{*}{\textcolor{tableMuted}{direction}}
  & \(\pi_0\) alone                       & \cellcolor{summaryband}45.12 & \textcolor{tableNegative}{\(-2.07\)} \\
  & \(t_{\mathrm{ref}}\) alone            & \cellcolor{summaryband}44.27 & \textcolor{tableNegative}{\(-2.92\)} \\
\addlinespace[3pt]\cmidrule(l){2-4}
\multirow[t]{2}{*}{\textcolor{tableMuted}{KL direction}}
  & reverse KL w. mass                            & \cellcolor{summaryband}45.36 & \textcolor{tableNegative}{\(-1.83\)} \\
  & reverse KL                 & \cellcolor{summaryband}44.36 & \textcolor{tableNegative}{\(-2.83\)} \\
\addlinespace[3pt]\cmidrule(l){2-4}
\textcolor{tableMuted}{extrapolation}
  & tilt off                             & \cellcolor{summaryband}34.22 & \textcolor{tableNegative}{\(-12.97\)} \\
\bottomrule
\end{tabularx}\arrayrulecolor{black}
\end{table}

We conduct ablations in Setting~A while keeping all other training and evaluation settings fixed.
Each block in Table~\ref{tab:ablation} removes or replaces one component of the full method: distance control, direction construction, the distillation loss, or extrapolation.

Disabling extrapolation causes the largest drop, reducing the average score from 47.19 to 34.22.
With extrapolation retained, removing the hard distance constraint lowers the score to 43.30, while removing only the soft penalty gives 45.59.
Thus, the hard bound contributes beyond the soft penalty.

Using only the student reference or teacher reference gives 45.12 and 44.27, respectively, whereas
combining the signals performs better.
Replacing forward KL with reverse KL lowers the score to 44.36.
Retaining the support-mass term improves this to 45.36, still below the full method.
This suggests that forward KL contributes beyond moving probability onto the candidate support.

\section{Conclusion}
\label{sec:conclusion}

We presented \ours{}, decoupling extrapolation direction from distance in extrapolative on-policy distillation.
The method combines two directional signals while enforcing a per-prefix KL bound on the target's distance from the teacher.
On three teacher--student settings, \ours{} outperforms OPD baselines on mathematical reasoning benchmarks.
Ablations show that combined direction signals, the distance constraint, and forward-KL distillation each contribute to these gains.

\FloatBarrier
\section*{Limitations}
Our experiments cover three teacher--student pairs from the Qwen3 family and
three mathematical-reasoning benchmarks. They do not establish whether the
same gains hold for other model families or tasks. The direction construction
uses a teacher reference checkpoint when one is available. Without that
checkpoint, the two directions coincide, so the method cannot use the teacher's
post-training shift as a separate signal. The distance bound and soft
extrapolation coefficient are hyperparameters; the method does not determine
an optimal extrapolation distance. The parameter study also shows that the
choice of distance bound matters across settings.

\section*{Ethical Considerations}
Distillation can transfer undesirable behavior from the teacher, including biased or harmful outputs.
Our mathematical-reasoning evaluations do not assess these behaviors or privacy risks.
Training and evaluation also incur energy costs.

\bibliography{refs,refs_extra}

\clearpage
\raggedbottom
\onecolumn
\appendix

\begingroup
\noindent\begin{minipage}[t]{\dimexpr(\textwidth-\columnsep)/2\relax}
\section{Training and evaluation details}
\label{app:hyperparams}

Our experiments focus on mathematical reasoning, with evaluation on English-language competition problems.
We do not assess performance across languages, specific linguistic phenomena, or demographic groups.

\paragraph{Training.}
All methods share the same default training setup of \citet{rethinking}, except that we set the maximum
response length to 4{,}096 rather than 7{,}168 tokens. This lies between the 3K and 7K moderate
lengths that they report as yielding the strongest results \citep{rethinking}.
Table~\ref{tab:hyperparams} lists the default training hyperparameters.

\end{minipage}\hfill
\begin{minipage}[t]{\dimexpr(\textwidth-\columnsep)/2\relax}
\paragraph{Evaluation.}
We report avg@16 using temperature 0.7, top-\(p\) sampling with \(p=0.95\), and a maximum response length of 31{,}744 tokens.
Responses are graded using the rule-based verifier of \citet{verl}. Each setting is evaluated with the same template used during training.

\paragraph{Hardware and compute.}
All methods were trained on AWS nodes with eight NVIDIA A100 40GB GPUs per node.
Settings A and B used one node, and Setting C used two nodes.
Training for the reported experiments required approximately 9{,}000 GPU-hours across 50 runs:
33 in Setting A, eight in B, and nine in C.
These runs cover the main comparisons, ablations, parameter sweeps, and constraint-activity analysis.
The calculation uses approximate average training times of 20, 16, and 18 hours per run in Settings A, B, and C, respectively.
\end{minipage}
\par
\endgroup

\par\medskip
\begingroup
\noindent\begin{minipage}{\textwidth}
\captionsetup{type=table,hypcap=false}
\centering
\small
\caption{Default training hyperparameters.}
\label{tab:hyperparams}
\begin{tabular*}{\textwidth}{@{\extracolsep{\fill}}lclc@{}}
\arrayrulecolor{tableNavy}\toprule[0.9pt]\arrayrulecolor{tableRule}
\textcolor{black}{\textbf{Item}} & \textcolor{black}{\textbf{Value}} & \textcolor{black}{\textbf{Item}} & \textcolor{black}{\textbf{Value}} \\
\midrule
Training temperature & 1.0 & Learning rate & 1e-6 \\
Global batch size & 64 & Epoch & 1 \\
Mini batch size & 64 & Actor KL-loss coefficient & 0.0 \\
Rollout number & 4 & Optimizer & AdamW \\
LogProb top-\(K\) & 16 & LR schedule & constant \\
Top-\(p\) & 1.0 & Gradient clipping & 1.0 \\
Max prompt length & 1024 & Loss aggregation & token-mean \\
Max response length & 4096 & Model dtype & fp32 \\
\bottomrule
\end{tabular*}\arrayrulecolor{black}
\end{minipage}
\par
\endgroup

\clearpage
\begingroup
\noindent\begin{minipage}[t]{\dimexpr(\textwidth-\columnsep)/2\relax}
\section{Baseline configurations}
\label{app:baseline-search}

OPD follows the implementation released by \citet{rethinking}.
ExOPD's reward-correction variant corresponds to the \(\lambda>1\) case of G-OPD with \(b=t_{\mathrm{ref}}\).
For baselines with method-specific hyperparameters, we use the authors' recommended values and also search their recommended ranges in Setting~A.

Table~\ref{tab:search} reports the baseline configurations evaluated in Setting~A. For each method, \(\dagger\) marks the authors' default configuration, which is used in Table~\ref{tab:settingA}; the remaining rows vary its method-specific parameters.
\end{minipage}\hfill
\begin{minipage}[t]{\dimexpr(\textwidth-\columnsep)/2\relax}
For EOPD, \(\tau\) is the teacher-entropy threshold and \(\alpha\) weights the additional forward-KL loss \citep{eopd}.
For TOP-D, \(\alpha\) is the teacher's weight in the proximal teacher--student mixture \citep{topd}.
For Direct-OPD, \(\alpha\) is the regularization coefficient introduced in Section~\ref{sec:prelim:gopd}; for ExOPD, \(\lambda\) is the reward-scaling coefficient.
The published configuration achieves the highest average score among the tested configurations for all four baselines.
\end{minipage}
\par
\endgroup
\par\medskip
\begingroup
\noindent\begin{minipage}{\textwidth}
\captionsetup{type=table,hypcap=false}
\centering
\small
\renewcommand{\arraystretch}{1.0}
\caption{Baseline parameter sensitivity in Setting~A. \(\dagger\) marks the authors' default configuration, which is used in Table~\ref{tab:settingA}.
Rows are ordered by \emph{Avg} within each method.}
\label{tab:search}
\begin{tabularx}{\textwidth}{lYNNNN}
\arrayrulecolor{tableNavy}\toprule[0.9pt]\arrayrulecolor{tableRule}
\rowcolor{white}
\textcolor{black}{\textbf{Method}} & \textcolor{black}{\textbf{Configuration}} & \textcolor{black}{\textbf{AIME24}} & \textcolor{black}{\textbf{AIME25}} & \textcolor{black}{\textbf{AMC23}} & \textcolor{black}{\textbf{Avg}} \\
\midrule
\multirow{4}{*}{\shortstack[l]{\textcolor{tableMuted}{EOPD}\\\citep{eopd}}}
  & \cellcolor{refband}\(\tau=0.8,\ \alpha=1.0\)\(^\dagger\) & \cellcolor{refband}20.00 & \cellcolor{refband}21.67 & \cellcolor{refband}54.22 & \cellcolor{summaryband}\best{31.96} \\
  & \(\tau=0.4,\ \alpha=1.0\)             & 22.29 & 19.38 & 54.06 & \cellcolor{summaryband}31.91 \\
  & \(\tau=0.8,\ \alpha=2.0\)             & 22.92 & 19.38 & 51.88 & \cellcolor{summaryband}31.39 \\
  & \(\tau=0.4,\ \alpha=2.0\)             & 22.29 & 15.83 & 50.94 & \cellcolor{summaryband}29.69 \\
\cmidrule(l){2-6}
\multirow{3}{*}{\shortstack[l]{\textcolor{tableMuted}{TOP-D}\\\citep{topd}}}
  & \cellcolor{refband}\(\alpha=0.2\)\(^\dagger\) & \cellcolor{refband}25.62 & \cellcolor{refband}23.54 & \cellcolor{refband}64.06 & \cellcolor{summaryband}\best{37.74} \\
  & \(\alpha=0.1\)                        & 22.71 & 20.62 & 64.53 & \cellcolor{summaryband}35.95 \\
  & \(\alpha=0.5\)                        & 23.54 & 16.25 & 62.66 & \cellcolor{summaryband}34.15 \\
\cmidrule(l){2-6}
\multirow{3}{*}{\shortstack[l]{\textcolor{tableMuted}{Direct-OPD}\\\citep{directopd}}}
  & \cellcolor{refband}\(\alpha=1.0\) adaptive\(^\dagger\) & \cellcolor{refband}23.75 & \cellcolor{refband}20.62 & \cellcolor{refband}57.97 & \cellcolor{summaryband}\best{34.11} \\
  & \(\alpha=0.01\) fixed                 & 23.96 & 20.21 & 52.50 & \cellcolor{summaryband}32.22 \\
  & \(\alpha=0.1\) fixed                  & 22.29 & 19.38 & 51.56 & \cellcolor{summaryband}31.08 \\
\cmidrule(l){2-6}
\multirow{3}{*}{\shortstack[l]{\textcolor{tableMuted}{ExOPD}\\\citep{exopd}}}
  & \cellcolor{refband}\(\lambda=1.25\)\(^\dagger\) & \cellcolor{refband}17.08 & \cellcolor{refband}24.79 & \cellcolor{refband}54.37 & \cellcolor{summaryband}\best{32.08} \\
  & \(\lambda=1.0\)                       & 15.83 & 17.71 & 44.69 & \cellcolor{summaryband}26.08 \\
  & \(\lambda=2.0\)                       & 17.50 & 13.75 & 38.75 & \cellcolor{summaryband}23.33 \\
\bottomrule
\end{tabularx}\arrayrulecolor{black}
\end{minipage}
\par
\endgroup

\clearpage
\begingroup
\noindent\begin{minipage}{\textwidth}
\noindent\begin{minipage}[t]{\dimexpr(\textwidth-\columnsep)/2\relax}
\section{Parameter sensitivity}
\label{app:sweeps}

\paragraph{Setting A.}
We evaluate \(\epsilon\in\{1,4,8\}\) and \(\kappa\in\{0.02,0.05,0.1\}\).
Table~\ref{tab:sweepA} reports the average scores, which range from 45.00 to 47.83.
Underlining denotes the configuration used in the main experiments.

\par\medskip\noindent
\begin{minipage}{\columnwidth}
\captionsetup{type=table}
\centering
\small
\renewcommand{\arraystretch}{1.15}
\caption{Setting~A parameter sensitivity. \textbf{Avg} over AIME24, AIME25, and AMC23 for each \((\epsilon,\kappa)\) pair.}
\label{tab:sweepA}
\begin{tabularx}{\linewidth}{lRRR}
\arrayrulecolor{tableNavy}\toprule[0.9pt]\arrayrulecolor{tableRule}
\rowcolor{white}
 & \textcolor{black}{\(\epsilon=1\)} & \textcolor{black}{\(\epsilon=4\)} & \textcolor{black}{\(\epsilon=8\)} \\
\midrule
\(\kappa=0.02\) & 46.53 & 45.00 & 46.55 \\
\(\kappa=0.05\) & \cellcolor{summaryband}\underline{47.19} & 47.83 & 45.45 \\
\(\kappa=0.10\) & 45.38 & 45.99 & 45.66 \\
\bottomrule
\end{tabularx}\arrayrulecolor{black}
\end{minipage}
\par\medskip

\paragraph{Setting C.}
We select \(\epsilon\) based on training-time constraint activity.
With \(\epsilon=1\), the constraint binds at about 89\% of positions early in training.
Increasing \(\epsilon\) to 4 reduces this to about 60\%, similar to Setting~A.
We therefore use \(\epsilon=4\) in the main experiments and report larger radii in Table~\ref{tab:sweepC}.
The average score increases from 43.21 at \(\epsilon=1\) to 48.45 at \(\epsilon=16\) over the tested values.

\par\medskip\noindent
\begin{minipage}{\columnwidth}
\captionsetup{type=table}
\centering
\small
\renewcommand{\arraystretch}{1.15}
\caption{Setting~C sensitivity to \(\epsilon\). \textbf{Avg} over AIME24, AIME25, and AMC23.}
\label{tab:sweepC}
\begin{tabularx}{\linewidth}{lRRRR}
\arrayrulecolor{tableNavy}\toprule[0.9pt]\arrayrulecolor{tableRule}
\rowcolor{white}
 & \textcolor{black}{\(\epsilon=1\)} & \textcolor{black}{\(\epsilon=4\)} & \textcolor{black}{\(\epsilon=8\)} & \textcolor{black}{\(\epsilon=16\)} \\
\midrule
\textbf{Avg} & 43.21 & \cellcolor{summaryband}\underline{46.27} & 47.05 & 48.45 \\
\bottomrule
\end{tabularx}\arrayrulecolor{black}
\end{minipage}
\par\medskip

\end{minipage}\hfill
\begin{minipage}[t]{\dimexpr(\textwidth-\columnsep)/2\relax}
\section{Per-benchmark ablations}
\label{app:ablations}

Table~\ref{tab:ablation-full} gives the AIME24, AIME25, and AMC23 scores for the ablations
summarized in Table~\ref{tab:ablation}.

\paragraph{Constraint variants.}
The \(-\kappa\) variant sets \(\kappa=0\) and retains \(\epsilon=1\).
The \(-\epsilon\) variant sets \(\epsilon=10^6\) to make the constraint slack and retains \(\kappa=0.05\).
The \(-\)both variant uses \(\kappa=0\) and \(\epsilon=10^6\), with the extrapolation coefficient limited only by the numerical cap of \(10^4\).

\paragraph{Direction variants.}
The \(\pi_0\)-alone and \(t_{\mathrm{ref}}\)-alone variants replace the combined score by \(\Delta_{\pi_0}^c(h,a)\) and \(\Delta_T^c(h,a)\), respectively, while retaining \(\epsilon=1\) and \(\kappa=0.05\).
If the selected signal's reference distance is at most \(10^{-3}\), its score is set to zero.

\paragraph{KL and target variants.}
Reverse KL uses \(D_{\mathrm{KL}}(\pi_{\theta,S}(\cdot\mid h)\|\tilde q(\cdot\mid h))\), with the student renormalized on \(S(h)\).
The variant with mass adds \(-\log\pi_\theta(S(h)\mid h)\), retaining the support-mass term in Eq.~\eqref{eq:fkl-support}.
Both variants retain the full method's target construction.
The tilt-off variant uses the renormalized teacher distribution \(q_S^+(\cdot\mid h)\) as the target and retains forward-KL distillation.

\end{minipage}
\par\medskip
\begingroup
\captionsetup{type=table,hypcap=false}
\centering
\small
\renewcommand{\arraystretch}{1.15}
\caption{Per-benchmark ablations in Setting~A, corresponding to Table~\ref{tab:ablation}. Each block removes or replaces one component of the full method;
\(\Delta\) denotes the change in \textbf{Avg} relative to the full method.}
\label{tab:ablation-full}
\begin{tabularx}{\textwidth}{llRRRRR}
\arrayrulecolor{tableNavy}\toprule[0.9pt]\arrayrulecolor{tableRule}
\rowcolor{white}
\textcolor{black}{\textbf{Component}} & \textcolor{black}{\textbf{Variant}} & \textcolor{black}{\textbf{AIME24}} & \textcolor{black}{\textbf{AIME25}} & \textcolor{black}{\textbf{AMC23}} & \textcolor{black}{\textbf{Avg}} & \textcolor{black}{\(\boldsymbol{\Delta}\)} \\
\midrule
\rowcolor{oursband}
\multicolumn{2}{l}{\cellcolor{oursband}full} & 35.62 & 35.00 & 70.94 & \cellcolor{ourssummary}\best{47.19} & --- \\
\midrule
\multirow[t]{3}{*}{\textcolor{tableMuted}{constraint}}
  & \(- \kappa\)            & 34.38 & 30.21 & 72.19 & \cellcolor{summaryband}45.59 & \textcolor{tableNegative}{\(-1.60\)} \\
  & \(-\epsilon\)                         & 30.21 & 26.88 & 72.81 & \cellcolor{summaryband}43.30 & \textcolor{tableNegative}{\(-3.89\)} \\
  & \(-\)both                            & 30.00 & 26.46 & 71.72 & \cellcolor{summaryband}42.73 & \textcolor{tableNegative}{\(-4.46\)} \\
\addlinespace[3pt]\cmidrule(l){2-7}
\multirow[t]{2}{*}{\textcolor{tableMuted}{direction}}
  & \(\pi_0\) alone                       & 31.67 & 30.42 & 73.28 & \cellcolor{summaryband}45.12 & \textcolor{tableNegative}{\(-2.07\)} \\
  & \(t_{\mathrm{ref}}\) alone            & 29.38 & 30.00 & 73.44 & \cellcolor{summaryband}44.27 & \textcolor{tableNegative}{\(-2.92\)} \\
\addlinespace[3pt]\cmidrule(l){2-7}
\multirow[t]{2}{*}{\textcolor{tableMuted}{KL direction}}
  & reverse KL w. mass                            & 32.08 & 33.54 & 70.47 & \cellcolor{summaryband}45.36 & \textcolor{tableNegative}{\(-1.83\)} \\
  & reverse KL                 & 32.71 & 28.96 & 71.41 & \cellcolor{summaryband}44.36 & \textcolor{tableNegative}{\(-2.83\)} \\
\addlinespace[3pt]\cmidrule(l){2-7}
\textcolor{tableMuted}{extrapolation}
  & tilt off                             & 25.00 & 18.75 & 58.91 & \cellcolor{summaryband}34.22 & \textcolor{tableNegative}{\(-12.97\)} \\
\bottomrule
\end{tabularx}\arrayrulecolor{black}
\endgroup

\end{minipage}
\par
\endgroup
\clearpage
\begingroup
\noindent\begin{minipage}{\textwidth}
\noindent\begin{minipage}[t]{\dimexpr(\textwidth-\columnsep)/2\relax}
\section{Additional analysis}
\label{app:additional-analysis}
\subsection{Response length}
\label{app:response-length}

Figure~\ref{fig:response-length-details} gives the full response-length comparison, including the initial students and teachers.
\end{minipage}\hfill
\begin{minipage}[t]{\dimexpr(\textwidth-\columnsep)/2\relax}
\subsection{Entropy gap}
\label{app:entropy-gap}

Figure~\ref{fig:entropy-gap} shows the student--teacher entropy gap across the three settings.
\ours{} has the largest mean student--teacher entropy gap \(H(\pi_\theta(\cdot\mid h))-H(q^+(\cdot\mid h))\) in all three settings.
Across the tested methods, a smaller mean entropy gap does not consistently correspond to a higher average score.
\end{minipage}
\par\medskip
\begingroup
\captionsetup{type=figure,hypcap=false}
\centering
\includegraphics[width=0.90\textwidth]{figures/length_grid}
\caption{\textbf{Longer responses do not explain the gain.}
\textbf{Top:} mean response length for each method; the dotted line marks the 31{,}744-token generation limit.
\textbf{Bottom:} average score versus mean response length for the trained methods.}
\label{fig:response-length-details}
\endgroup

\par\medskip
\begingroup
\captionsetup{type=figure,hypcap=false}
\centering
\newcommand{\entropykey}[2]{\textcolor[HTML]{#1}{\rule[0.5ex]{1.2em}{0.8pt}}\,#2}
{\footnotesize
\entropykey{0072B2}{\ours{}}\quad
\entropykey{D55E00}{OPD}\quad
\entropykey{E69F00}{Direct-OPD}\quad
\entropykey{56B4E9}{TOP-D}\quad
\entropykey{009E73}{EOPD}\quad
\entropykey{CC79A7}{ExOPD}\quad
\entropykey{666666}{SFT + OPD}\par}
\smallskip
\hspace*{0.05\textwidth}%
\makebox[0.316\textwidth]{\small Setting A}%
\makebox[0.316\textwidth]{\small Setting B}%
\makebox[0.316\textwidth]{\small Setting C}\par
\includegraphics[width=0.90\textwidth,trim=0 0 0 213.27952bp,clip]{figures/entropy_grid}
\caption{\textbf{Student--teacher entropy gaps.} For a full-vocabulary token distribution \(p(\cdot\mid h)\), its entropy is \(H(p(\cdot\mid h))=-\sum_a p(a\mid h)\log p(a\mid h)\).
\textbf{Top:} \(H(\pi_\theta(\cdot\mid h))-H(q^+(\cdot\mid h))\) over training; values below zero indicate that the student has lower entropy than the teacher. Faint lines show per-step values and solid lines show a 9-step rolling mean.
\textbf{Bottom:} average score versus the mean entropy gap.
Entropies are measured on each method's own on-policy prefixes, rather than on a common prefix set.}
\label{fig:entropy-gap}
\endgroup

\end{minipage}
\par
\endgroup
\clearpage
\twocolumn
\section{Computational-cost experiment setup}
\label{app:computational-cost}

We use the model checkpoints from Setting~A on two NVIDIA A100 40GB GPUs on one node.
All methods process the same saved batch of 16 prompts with four responses per prompt.
The maximum prompt and response lengths are 1{,}024 and 4{,}096 tokens, respectively.

Timing a full training step would include differences in the responses generated by each method.
In particular, response length affects both generation and training time.
Reusing the same saved responses fixes the token sequences and their lengths, allowing us to compare the cost of computing supervision and updating the student.

We discard the first two training steps as warm-up and report the mean time over the following three steps.
The timed computation includes model scoring, target or reward construction, and student optimization.
Throughput counts valid prompt and response tokens across both GPUs.

\section{Extrapolative distillation objectives}
\label{app:prior-objectives}

This appendix gives the baseline objectives summarized in Section~\ref{sec:prelim}.
\paragraph{ExOPD.}
For an arbitrary reference \(b\), \citet{exopd} rewrite reverse-KL OPD as
\begin{equation}
\label{eq:gopd-rewrite}
\begin{aligned}
&\max_\theta\;\underset{\substack{x\sim\mathcal D\\y\sim\pi_\theta(\cdot\mid x)}}{\mathbb E}
\Bigl[\log\frac{q^+(y\mid x)}{b(y\mid x)}\\
&-D_{\mathrm{KL}}\!\left(\pi_\theta(\cdot\mid x)\,\|\,b(\cdot\mid x)\right)\Bigr].
\end{aligned}
\end{equation}
Scaling the reward by \(\lambda>0\) gives Eq.~\eqref{eq:gopd-opt}.
The target interpolates between \(b(\cdot\mid x)\) and \(q^+(\cdot\mid x)\) for \(\lambda<1\), recovers \(q^+(\cdot\mid x)\)
at \(\lambda=1\), and continues past it along \(\log(q^+(y\mid x)/b(y\mid x))\) for \(\lambda>1\).

\paragraph{Direct-OPD.}
Direct-OPD \citep{directopd} transfers the teacher's pre- to post-training log-ratio
and regularizes the student toward its own initialization:
\begin{equation}
\label{eq:dopd-obj}
\begin{aligned}
&\max_\theta\;\underset{\substack{x\sim\mathcal D\\y\sim\pi_\theta(\cdot\mid x)}}{\mathbb E}
\Bigl[\log\frac{q^+(y\mid x)}{t_{\mathrm{ref}}(y\mid x)}\\
&-\alpha D_{\mathrm{KL}}\!\left(\pi_\theta(\cdot\mid x)\,\|\,\pi_0(\cdot\mid x)\right)\Bigr].
\end{aligned}
\end{equation}
Its optimum is Eq.~\eqref{eq:dopd-opt}.
Direct-OPD adapts a shared regularization coefficient using the batch's mean teacher-shift reward.
For the corresponding fixed-prefix objective, denote its target by \(q_{\mathrm{Direct}}(\cdot\mid h)\). It can be written relative to the teacher as
\begin{equation}
\begin{aligned}
&q_{\mathrm{Direct}}(a\mid h)
\propto\pi_0(a\mid h)\exp[\Delta_T(h,a)/\alpha]\\
&=q^+(a\mid h)\exp\!\bigl[\alpha^{-1}\Delta_T(h,a)-\Delta_{\pi_0}(h,a)\bigr].
\end{aligned}
\end{equation}
Changing \(\alpha\) generally changes both direction and distance relative to \(q^+(\cdot\mid h)\).
ExOPD with a fixed reference keeps the direction fixed, but the realized distance still depends on the scale of its log-ratio at each prefix.
Our method constructs the direction first, then selects a separate coefficient at each prefix to enforce \(D_{\mathrm{KL}}(\tilde q(\cdot\mid h)\|q_S^+(\cdot\mid h))\le\epsilon\).

\section{Direction and distance in target construction}
\label{app:theory}

We provide proofs for Propositions~\ref{prop:theory:rare-errors} and~\ref{prop:theory:relative-direction}, together with examples and further discussion of direction and distance in target construction.
We fix a prefix \(h\) and use the finite candidate support \(S(h)\).
The renormalized teacher \(q_S^+(\cdot\mid h)\) is positive on this support, and candidate targets are zero outside it.

\subsection{Distance}
\label{app:theory-distance}

\paragraph{Proof of Proposition~\ref{prop:theory:rare-errors}.}
Let \(B\) and \(\rho(h)=q_S^+(B\mid h)\) be as in the proposition.
Let \(w(h)=q(B\mid h)\) and let \(H_{\mathrm{bin}}(w)=-w\log w-(1-w)\log(1-w)\) denote binary entropy.
Grouping tokens into \(B\) and its complement gives
\begin{equation}
\begin{aligned}
&D_{\mathrm{KL}}(q(\cdot\mid h)\|q_S^+(\cdot\mid h))\\
&\quad\ge w(h)\log\frac{w(h)}{\rho(h)}
+(1-w(h))\log\frac{1-w(h)}{1-\rho(h)}\\
&\quad=w(h)\log\frac1{\rho(h)}
+(1-w(h))\log\frac1{1-\rho(h)}\\
&\qquad-H_{\mathrm{bin}}(w(h))\\
&\quad\ge w(h)\log\frac1{\rho(h)}-\log2.
\end{aligned}
\end{equation}
Combining this inequality with the KL constraint proves the bound. \hfill\(\square\)

\paragraph{Fixed-coefficient counterexample.}
For every fixed extrapolation coefficient \(\beta>0\) and target mass \(m\in(0,1)\), there are centered score errors \(e_\rho(h,a)\) with \(\mathbb E_{a\sim q_S^+(\cdot\mid h)}[e_\rho(h,a)^2]\to0\) such that the normalized target \(q_\beta(a\mid h)\propto q_S^+(a\mid h)\exp[\beta e_\rho(h,a)]\) assigns probability \(m\) to \(B\), even as \(\rho(h)\to0\).

For the fixed-coefficient construction, let \(\mathbf1_B(a)\) be the indicator of membership in \(B\), and set \(c_\rho(h)=\beta^{-1}\log\!\bigl(m(1-\rho(h))/((1-m)\rho(h))\bigr)\). Starting from a zero score, introduce the error \(e_\rho(h,a)=c_\rho(h)(\mathbf1_B(a)-\rho(h))\).
This error is centered, and
\begin{equation}
\begin{aligned}
&\mathbb E_{a\sim q_S^+(\cdot\mid h)}[e_\rho(h,a)^2]\\
&=\frac{\rho(h)(1-\rho(h))}{\beta^2}
\left(\log\frac{m(1-\rho(h))}{(1-m)\rho(h)}\right)^2\to0.
\end{aligned}
\end{equation}
The score difference between \(B\) and its complement is \(c_\rho(h)\).
Therefore,
\begin{equation}
q_\beta(B\mid h)=\frac{\rho(h)e^{\beta c_\rho(h)}}{1-\rho(h)+\rho(h)e^{\beta c_\rho(h)}}=m.
\end{equation}
This establishes the contrast. \hfill\(\square\)

\paragraph{Adaptation to score scale.}
For an extrapolation coefficient \(\beta\ge0\), consider the normalized target \(\tilde q_\beta(a\mid h)\propto q_S^+(a\mid h)\exp[\beta d_\eta(h,a)]\).
For a scale factor \(c>0\), if the constraint binds, replacing \(d_\eta(h,a)\) by \(c d_\eta(h,a)\) leaves the boundary target unchanged whenever the coefficient cap permits reaching it.
The selected coefficient becomes \(\beta/c\), and
\begin{equation}
\begin{aligned}
&q_S^+(a\mid h)\exp\!\left[\frac{\beta}{c}\,c d_\eta(h,a)\right]\\
&\quad=q_S^+(a\mid h)\exp[\beta d_\eta(h,a)].
\end{aligned}
\end{equation}
Thus, at active prefixes, a positive rescaling of the direction changes the selected coefficient but not the target.

\paragraph{Connection to response length.}
ExOPD reports increasing response length under stronger extrapolation and identifies amplification of biased log-ratios as a possible source of instability \citep{exopd}.
The fixed-coefficient counterexample above describes one mechanism that could contribute to this behavior.
If teacher-rare tokens that prolong or repeat reasoning receive spuriously large scores, fixed-strength extrapolation can assign them substantial probability.
Our KL constraint limits this amplification at each prefix.

\subsection{Direction}
\label{app:theory-direction}

\paragraph{Proof of Proposition~\ref{prop:theory:relative-direction}.}
Write the objective in Eq.~\eqref{eq:relative-direction} as \(F_h(d(h,\cdot))\).
Completing the square gives
\begin{equation}
\begin{aligned}
&F_h(d(h,\cdot))-F_h(d_\eta(h,\cdot))\\
&\quad=\left(\frac{1}{\sigma_T^2(h)}+\frac{1}{\sigma_\pi^2(h)}\right)\\
&\qquad\cdot\mathbb E_{a\sim q_S^+(\cdot\mid h)}[(d(h,a)-d_\eta(h,a))^2].
\end{aligned}
\end{equation}
The solution in Eq.~\eqref{eq:direction-implementation} is centered, so it is feasible.
The coefficient is positive, establishing uniqueness. \hfill\(\square\)

This solution gives more weight to the nearer reference.
Its direction is proportional to \(\sigma_\pi(h)u(h,a)+\sigma_T(h)v(h,a)\), where \(u(h,a)=\Delta_T^c(h,a)/\sigma_T(h)\) and \(v(h,a)=\Delta_{\pi_0}^c(h,a)/\sigma_\pi(h)\).

\paragraph{Conflicting signals.}
If the normalized signals point in opposite directions, \(u(h,a)=-v(h,a)\) for all \(a\in S(h)\), their equal-weight sum vanishes.
Our combination instead satisfies
\begin{equation}
\label{eq:theory:opposite-signals}
d_\eta(h,a)=\frac{\sigma_T(h)\sigma_\pi(h)(\sigma_\pi(h)-\sigma_T(h))u(h,a)}
{\sigma_T^2(h)+\sigma_\pi^2(h)}.
\end{equation}
It remains nonzero when the reference distances differ and follows the signal from the nearer reference.
When the reference distances are equal, the signals cancel and the target remains \(q_S^+(\cdot\mid h)\).

For example, consider a two-token support with \(\pi_0(\cdot\mid h)=(0.1,0.9)\), \(t_{\mathrm{ref}}(\cdot\mid h)=(0.5,0.5)\), and \(q^+(\cdot\mid h)=q_S^+(\cdot\mid h)=(0.3,0.7)\).
The teacher's post-training shift favors the second token, while the teacher--student gap favors the first.
Their normalized directions are opposite.
Our combination gives \(d_\eta(h,\cdot)\approx(-0.16,0.07)\), so for an extrapolation coefficient \(\beta\ge0\), the target \(\tilde q_\beta(a\mid h)\propto q_S^+(a\mid h)\exp[\beta d_\eta(h,a)]\) moves probability toward the second token.

Suppose the desired distribution is \(\pi^\star(\cdot\mid h)=(0.2,0.8)\), and assess a policy \(\pi(\cdot\mid h)\) by \(-D_{\mathrm{KL}}(\pi^\star(\cdot\mid h)\|\pi(\cdot\mid h))\).
The teacher is better than the initial student: their divergences from \(\pi^\star(\cdot\mid h)\) are approximately \(0.0257\) and \(0.0444\), respectively.
Along the constructed direction, \(\beta\approx2.38\) gives \(\tilde q_\beta(\cdot\mid h)\approx\pi^\star(\cdot\mid h)\).

\end{document}